\documentclass{article}

\usepackage[preprint]{neurips_2019}

\usepackage[utf8]{inputenc} % allow utf-8 input
\usepackage[T1]{fontenc}    % use 8-bit T1 fonts
\usepackage{hyperref}       % hyperlinks
\usepackage{url}            % simple URL typesetting
\usepackage{booktabs}       % professional-quality tables
\usepackage{amsfonts}       % blackboard math symbols
\usepackage{nicefrac}       % compact symbols for 1/2, etc.
\usepackage{microtype}      % microtypography
\usepackage{graphicx}
\usepackage{amsmath}
\usepackage{caption}
\usepackage[table,xcdraw]{xcolor}

\title{NASS-AI: Towards Digitization of Parliamentary Bills using Document Level Embedding and Bidirectional Long Short-Term Memory}

\author{%
  Adewale Akinfaderin\thanks{Equal Contribution}\\
  Data Duality Labs\\
  \texttt{aaa12g@my.fsu.edu} \\
 \And
   Olamilekan Wahab\footnotemark[1]  \\
   Independent Researcher \\
   \texttt{olamyy53@gmail.com} \\
}
\begin{document}

\maketitle
\begin{abstract}
  There has been several reports in the Nigerian and International media about the Senators and House of Representative Members of the Nigerian National Assembly (NASS) being the highest paid in the world. Despite this high-level of parliamentary compensation and a lack of  oversight, most of the legislative duties like bills introduced and vote proceedings are shrouded in mystery without an open and annotated corpus. In this paper, we present results from ongoing research on the categorization of bills introduced in the Nigerian parliament since the fourth republic (1999 - 2018). For this task, we employed a multi-step approach which involves extracting text from scanned and embedded pdfs with low to medium quality using Optical Character Recognition (OCR) tools and labeling them into eight categories. We investigate the performance of document level embedding for feature representation of the extracted texts before using a Bidirectional Long Short-Term Memory (Bi-LSTM) for our classifier. The performance was further compared with other feature representation and machine learning techniques. We believe that these results are well-positioned to have a substantial impact on the quest to meet the basic open data charter principles.
\end{abstract}

\section{Introduction}
Given the challenges and precariousness facing developing and underdeveloped countries, the quality of policy-making and legislation is of enormous importance. This legislation can be used to impact the success of some of the United Nations Sustainable Development Goals (SDGs) like poverty alleviation, good public health system, quality education, economic growth and sustainability. Targets 16.6 and 16.7 from the UN SDGs are to \emph{"develop effective, accountable, and transparent institutions at all levels"} and to \emph{"ensure responsive, inclusive, participatory and representative decision making at all levels"} [1]. For countries in Sub-Saharan Africa to meet this target, an open data revolution needs to happen at all levels of government and more importantly, at the parliamentary level.

According to the Open Data Barometer by the World Wide Web Foundation, Nigeria is currently ranked 70th in the world with a score of 21/100 on open data initiatives based on readiness, implementation and impact [2]. To make the processing of creation, introduction and passage of parliamentary bills a force for public accountability, the information needs to be easier to analyze and process by the average citizen. This is not the case for the Nigerian parliamentary bill. In this work, we present a method to overcome some of these barriers. We propose to apply natural language processing techniques and machine learning algorithms to classify the bills.

Our dataset consists of 2,397 parliamentary bills introduced in the Nigerian National Assembly in the 4th republic (1999 - 2018). The models used in this project leverage recent advancements in information retrieval and text classification from three areas: optical character recognition (OCR), document level embedding and deep learning for text classification. We used OCR for text retrieval, Doc2Vec embedding model for feature representation and a Bidirectional Long Short-Term Memory (Bi-LSTM) network architecture for classification. Our model was able to accurately predict the different bill classes with a weighted average F1-score of 72\%.

\section{Related Work}
\paragraph{Classification of Legal Texts:}
In most developed countries, legal texts, parliamentary bills and court documents are available as structured corpus[3-4]. However, due to inefficiency and a lack of a proper data management system, these structured corpus are difficult to get in underdeveloped and some developing countries. Researchers in developing countries like Brazil recently used machine learning to classify legal documents to enable them to unclog the overloaded judicial system [5]. 
\parskip -5pt
\paragraph{Analysis of Parliamentary Votes and Proceedings:}
For a thorough understanding of government processes, analyzing patterns of legislative data is important. Apart from parliamentary bills, data from roll calls and votes can help in the interpretation of legislative actions [6-8]. Extensive work has been done on speeches in parliamentary debates with promising results [9]. Using the data from introduced bills vs passed bills, machine learning algorithms with word vectors representation have been used to forecast the probability of bills becoming  law in developed nations [10].
% \parskip -5pt
% \paragraph{Deep Learning for Document Classification:}
% Recent advances have shown how the combination of pre-trained semantic embeddings together with deep learning algorithms can be used to model data for various document classification tasks [11-12]. The natural language processing community is witnessing an incredible growth in pre-trained language representation models. However, image-based sequence system for text recognition are yet to achieve incredible results for real-world applications.
\parskip 2pt

\begin{figure}
  \centering
   \includegraphics[width=1\linewidth]{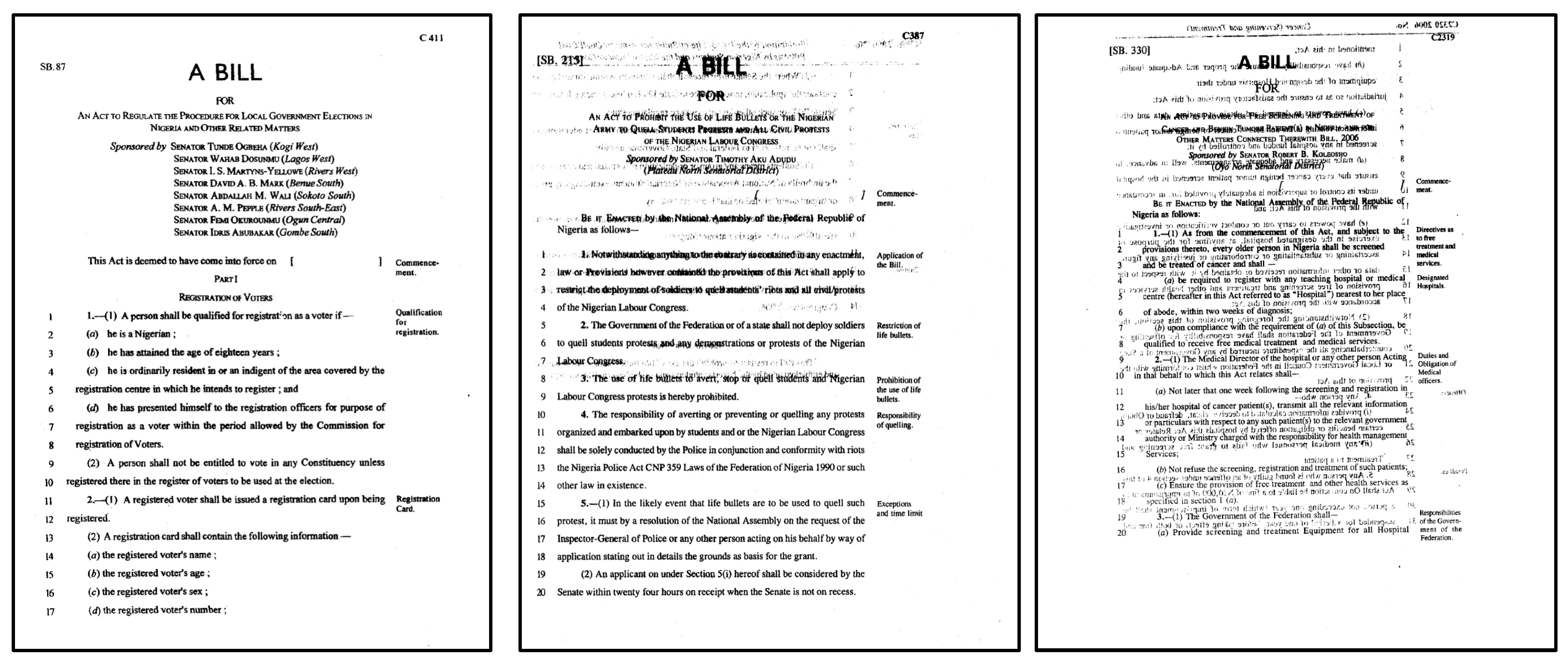}
  \caption{Three different bills showing some of the challenging quality of our parliamentary bills. Left: a bill to regulate local government elections. Center: a bill to prohibit the use of life bullets or Nigerian army to quell civil protests. Right: a bill to provide free screening and treatment of cancer and brain tumor. }
\end{figure}

\section{Study Methodology}

\subsection{Dataset, Labeling and Problem Definition}
Our dataset was scraped from the Nigeria National Assembly website [11], using the BeautifulSoup library [12]. The bills consist of two types of PDFs: embedded PDFs with good quality, and scanned PDFs with medium to low quality. The nature of the data brought up several challenges that forced our OCR choice. For the PDFS with embedded texts, we used PDFMiner to extract the texts [13]. For scanned PDFs, we converted each page of the PDF using pdf2image before using PyTesseract to extract the text from the image [14]. If the PDF has more than one page, the result from each image representing each page of the PDF is then concatenated to get the final output. If the PDF has one page, we get just a single extract as the output.

Each bill was hand-labeled into eight different categories. These eight classes are chosen as the policy area term to enable us to capture a short but accurate representation of the different facets of governance and the most important ministries. The eight classification groups shown with the frequency of the classes in Figure 2 are: laws, civil rights, safety and security; education, research and technology; government operation and international affairs; health and agriculture; labour, sports and social welfare; trade, commerce and macroeconomics; energy, environment and natural resources; public land, housing and transportation.
\begin{figure}
  \centering
   \includegraphics[width=0.95\linewidth]{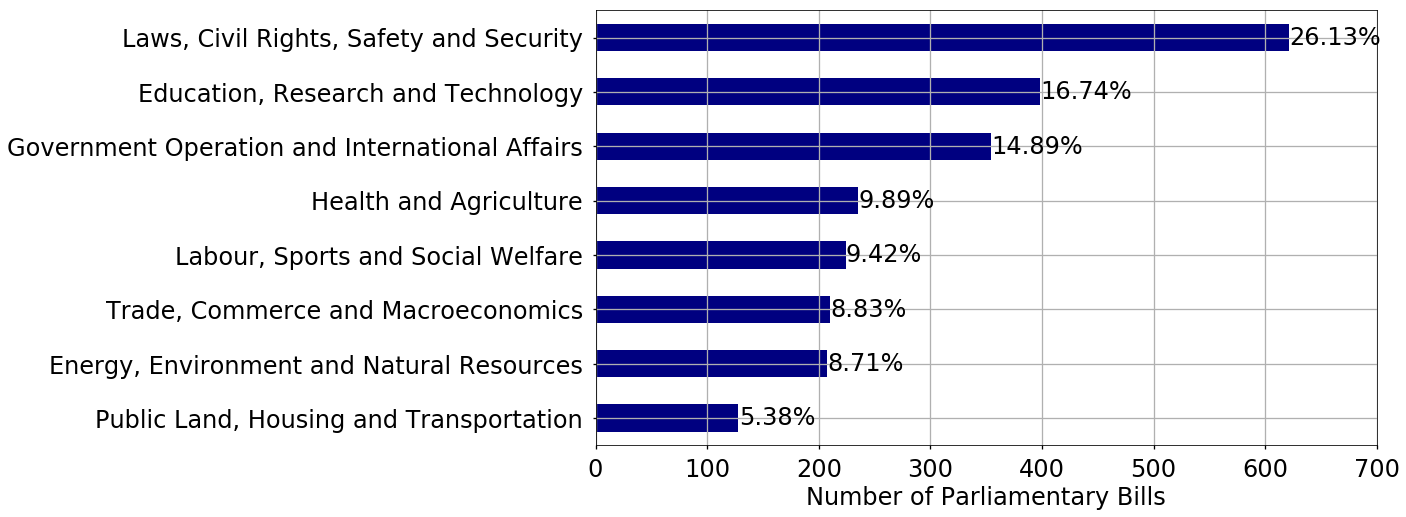}
  \caption{Distribution of parliamentary bills for the different classes. The ratio of the number of bills represented by each class to the total number of bills (2397) is shown on the bar chart. }
\end{figure}
\section{Experiments and Results}

For our text pre-processing, we used the NLTK library. First, we transformed the characters to lowercase. Then we removed unnecessary punctuation such as semicolon, full stop and colons. Next, we tokenize to split the text in each document into a list of words before lemmatizing to resolve words into their dictionary or canonical form. We used a maximum token number of 1,500 for our input layer. Our model consists of a Doc2Vec PV-DBoW embedding layer, a stacked Bi-LSTM layer with forward and backward LSTMs and two dense layers. The dimensions of our document embedding is $d = 400$. To make sure we are consistent with the dimensions, we used the same number of neurons for both LSTMs in the Bi-LSTM later, $n = 128$. The concatenated output from the Bi-LSTM layer is fed into a dense layer with 400 hidden neurons. We used the ReLu activation function for this layer. In order to combat overfitting, a dropout rate of 0.2 and a recurrent dropout rate of 0.2 were used. Unlike dropout where the input units are dropped, recurrent dropout drops the connection between recurrent units. The output from this dense layer is then fed into a second and final dense layer that has 8 hidden neurons, with a softmax activation function. These eight neurons corresponds to the eight different classes created for the dataset.

For evaluating the multi-label classification performance, we used the F1-score. This approach computes the metrics (precision and recall) independently for each class before taking the harmonic mean. This makes the metric less susceptible to imbalance class, unlike the classification accuracy. The F1-score for each class is a comprehensive measure of the model's accuracy.
\begin{align}
    F1-Score = \frac{2 \times Precision \times Recall}{Precision + Recall}
\end{align}
We trained and evaluated our deep learning model on Keras, with TensorFlow as the backend. For our experiment, we split the dataset into three parts: 1,509 bills in the training set, 377 bills in the validation set and 472 bills in the test set. Our model was trained on the training set, then the validation set was used to tune the model's hyperparameters and we used the test set for evaluating our model performance. For training, we used a mini-batch size of 256 and to learn the neural network model parameters, the categorical-cross entropy loss function and a gradient descent algorithm method based on Adaptive Moment Estimation (ADAM) were used. For ADAM, we used the following hyperparameters: learning rate, $\alpha = 0.001$; exponential decay rate for the first moment estimates, $\beta_1 = 0.9$, exponential decay rate for the second moment estimates $\beta_2 = 0.999$; and the hyperparameter that prevents division by zero, $\epsilon = 1 \times 10^{-8}$.
\begin{figure}[!h]
  \centering
   \includegraphics[width=1\linewidth]{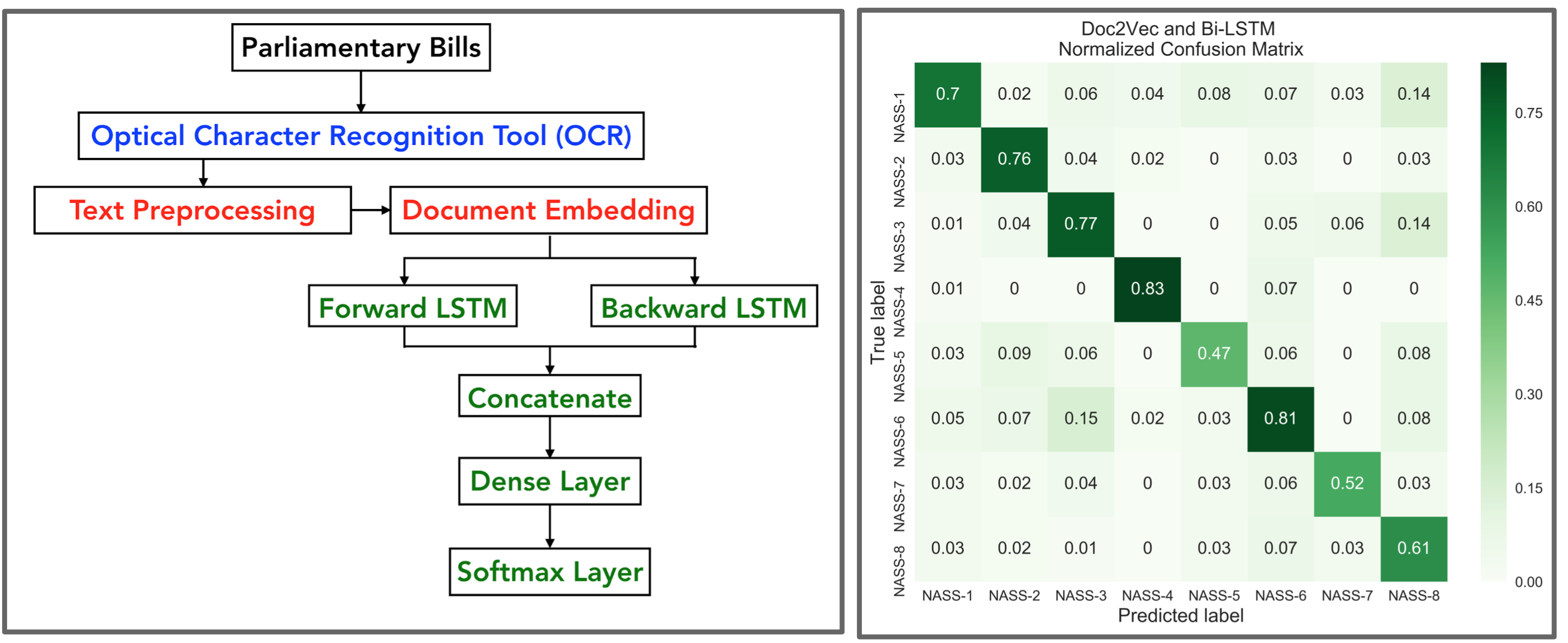}
  \caption{A schematic of our end-to-end architecture (Left) and the normalized confusion matrix in form of heatmap for model prediction on test data. (Right).}
\end{figure}
\begin{table}[!h]
\centering
\begin{tabular}{c|c|c|c|c}
\hline
\textbf{ID} & \textbf{Label Name}                                           & \textbf{Precision} & \textbf{Recall} & \textbf{F1-Score} \\ \hline
NASS-1      & \textit{Education, Research and Technology}              & 0.80               & 0.70            & 0.74              \\ \hline
NASS-2      & \textit{Energy, Environment and Natural Resources}       & 0.74               & 0.76           & 0.75              \\ \hline
NASS-3      & \textit{Government Operations and International Affairs} & 0.68               & 0.77            & 0.72              \\ \hline
NASS-4      & \textit{Health and Agriculture}                          & 0.91               & 0.83            & 0.87              \\ \hline
NASS-5      & \textit{Labour, Sports and Social Welfare}               & 0.75               & 0.47            & 0.58              \\ \hline
NASS-6      & \textit{Laws, Civil Rights, Safety and Security}         & 0.67               & 0.81            & 0.73              \\ \hline
NASS-7      & \textit{Public Land, Housing and Transportation}         & 0.80               & 0.52           & 0.63              \\ \hline
NASS-8      & \textit{Trade, Commerce and Macroeconomics}              & 0.55               & 0.61            & 0.58              \\ \hline
            & \textbf{Average}                                           & \textbf{0.73}      & \textbf{0.71}   & \textbf{0.72}     \\ \hline
\end{tabular}
\caption {Precision, Recall and F1-Score from Doc2Vec and Bi-LSTM for each class. }
\end{table}
Figure 3a highlights the precision, recall and F1 score for each class from our experiment. We also presented the confusion matrix that shows the actual and predicted labels in Figure 3b. Our model performed very well on Health and Agriculture bill category but least well on the Labour, Sports and Social Welfare bill category. The overall macro-average F1 score from our evaluation on the test set is 72\%. We also performed several experiments on our dataset with other feature representation methods and machine learning algorithms. The table and figure containing these results is presented in Appendix B. Our experiments shows that using document level embedding with a bidirectional LSTM network-based classifier outperform other baseline methods.

\section{Conclusion and Future Work}

In this work, we have presented a deep learning approach for classifying parliamentary bills in a low resource country. We explored optical character recognition (OCR) tools for converting low to medium quality PDF documents  and used document level embeddings to represent the output before feeding into a bi-directional LSTM model for classification performance. The experimental results shows that our model is effective for categorizing the bills will aid our large scale digitization efforts. However, we identified a key remaining challenge based on our results. The output from the pre-trained OCR tool is not generally a very accurate representation of the text in the bills, especially for the low-quality PDFs. A fascinating possibility is to solve this by training our custom OCR. The intensive acceleration of text detection research with novel deep learning methods can help us in this area. Methods such as region-based or single-shot based detectors can be employed. In addition to this, we can use image augmentation to alter the size, background noise or color of the bills. A large scale annotation effort of the texts can be as the labels for us to train our custom OCR for text identification and named entity recognition.

In summary, results that lead to accurate categorization of parliamentary bills are well-positioned to have substantial impact on governmental policies and on the quest for governments in low resource countries to meet the open data charter principles and United Nation's sustainability development goals on open government. Also, it can empower policymakers, stakeholders and governmental institutions to identify and monitor bills introduced to the National Assembly for research purposes and facilitate the efficiency of bill creation and open data initiatives. The next step in this project is to extend our method to other countries in Sub-Saharan Africa. We plan to design a tool that combines information from all bills and categories and make them easily accessible to everyone.
\section*{Acknowledgment}
We acknowledge Remi Akinfaderin and Wuraola Oyewusi for helping out with labeling the data. We also thank Busola Sanusi, Allen Akinkunle and Farouq Oyebiyi for the useful discussion and feedback.

\section*{References}

\parskip 5pt
[1] United Nations Sustainable Development Goals. https://sustainabledevelopment.un.org/sdg16.

[2] Open Data Barometer by the World Wide Web Foundation. https://opendatabarometer.org/4thedition/regional-snapshot/sub-saharan-africa/.

[3] Kalouli, A-L., Vrana, L., Fabella, V. M., Bellani, L.\ \& Hautli-Janisz, A.\ (2018) CoUSBi: A Structured and Visualized Legal Corpus of US State Bills. {\it Proceedings of the LREC 2018 “Workshop on Language Resources and Technologies for the Legal Knowledge Graph”},  Miyazaki, Japan.

[4] Yano, T., Smith, N. A. \ \& Wilkerson, J. D.\ (2012). CoUSBi: Textual Predictors of Bill Survival in Congressional Committees. {\it Proceedings of the Conference of the North American Chapter of the Association for Computational Linguistics (NAACL 2012)}, Montréal, Québec.

[5] Braz, F. A., Silva N. C., de Campos T. E. et al. (2018). Document classification using a Bi-LSTM to unclog Brazil’s supreme court. {\it NIPS 2018 Workshop on Machine Learning for the Developing World (ML4D)}, Montréal, Québec.

[6] Mickevicius, V., Krilavicius, T. \& Morkevicius, V. (2015). Classification of Short Legal Lithuanian Texts. {\it Proceedings of the 5th Workshop on Balto-Slavic Natural Language Processing}, pages 106–111, Hissar, Bulgaria.

[7] Gerrish, S. M., \& Blei, D. M. (2011).Predicting Legislative Roll Calls from Text. {\it Proceedings of the 28th International Conference on
Machine Learning}, Bellevue, WA, USA.

[8] Kraft, P. E.,  Jain, H., \& Rush, A. M. (2016). An Embedding Model for Predicting Roll-Call Votes. {\it Proceedings of the 2016 Conference on Empirical Methods in Natural Language Processing}, pages 2066–2070, Austin, TX, USA.

[9] Rohit, S. V. K. \& Singh, N. (2018). Analysis of Speeches in Indian Parliamentary Debates. {\it Proceedings of the 19th International Conference on Computational Linguistics and Intelligent Text Processing}, Hanoi, Vietnam.

[10] Nay, J. J. (2017). Predicting and understanding law-making with word vectors and an ensemble model. {\it PLoS ONE}. 12(5): e0176999.

% [11] Grnarova, P., Schmidt, F., Hyland, S. L. \& Eickhoff, C. (2016). Neural Document Embeddings for Intensive Care Patient Mortality Prediction. {\it NIPS 2016 Workshop on Machine Learning for Health.}, Barcelona, Spain.

% [12] Kim, D., Seo, D., Cho, S. \& Kang, P. (2019). Multi-co-training for document classification using various document representations: TF–IDF, LDA, and Doc2Vec. {\it Information Sciences}, Volume 477, Pages 15-29.

[11] National Assembly, Federal Republic of Nigeria. http://www.nassnig.org/document/bills.

[12] Richardson, L. (2007). {\it Beautiful Soup Documentation}.
2007

[13] Shinyama, Y. (2010).{\it PDFMiner: Python PDF parser and analyzer}.

[14] Smith, R. (2007). An overview of the Tesseract OCR engine.{\it Ninth International Conference on Document Analysis and Recognition (ICDAR 2007)}. Volume 2, pages 629–633. IEEE.

% [17] Le, Q. \& Mikolov, T. (2014). Distributed representations of sentences and documents. {\it In Proceedings of the 31st International Conference on
% Machine Learning (ICML 2014)}. Pages 1188–1196, Beijing, China.

[15] Hochreiter, S. \& Schmidhuber, J. (1997). Long short-term memory.
{\it Neural Computation}. 9(8):1735–1780.

[16] Goodfellow, I., Bengio, Y. \& Courville, A. (2016). Deep Learning. {\it MIT Press}.

[17] Graves, A. \& Schmidhuber, J.(2005). Framewise phoneme classification with bidirectional LSTM and other neural network architectures. {\it Neural Networks}. , 2005.

\newpage
\section*{Appendix A: Doc2Vec Embedding and Bi-LSTM Network Architecture}

Doc2Vec was created to extend embedding on Word2Vec. There are two different approaches for doc2vec: paragraph vector with distributed bags of words (PV-DBoW) and with distributed memory (PV-DM). 
The embedding used in this work is based on Doc2Vec PV-DBoW. In DBoW, the input is a vector representing the document and the order in which the words in the documents are arranged is ignored. The figure below shows the schematic of the PV-DBoW and Bi-LSTM network architecture.

Long short-term memory (LSTM) is a type of RNN based on gated recurrent unit [15-16]. This sequence model is more effective because it can learn long-term dependencies due to the introduction of recurrent network "cell" and gating mechanism. Given a current input $\mathbf{x}_{t}$, previous hidden state $\mathbf{h}_{t-1}$, and a previous cell state $\mathbf{c}_{t-1}$, an LSTM cell can compute current hidden state $\mathbf{h}_{t}$. The equations for the input gate $\mathbf{i}_{t}$, forget gate $\mathbf{f}_{t}$, output gate $\mathbf{o}_{t}$, memory cell $\mathbf{c}_{t}$ and current hidden state $\mathbf{h}_{t}$ are:
\begin{align*} 
    \mathbf{i}_{t} &=\sigma\left(\mathbf{W}_{i}[\mathbf{x}_{t}, \mathbf{h}_{t-1},\mathbf{c}_{t-1}]+\mathbf{b}_{i}\right) \\ 
    \mathbf{f}_{t} &=\sigma\left(\mathbf{W}_{f}[\mathbf{x}_{t}, \mathbf{h}_{t-1},\mathbf{c}_{t-1}]+\mathbf{b}_{f}\right) \\ 
    \mathbf{o}_{t} &=\sigma\left(\mathbf{W}_{o}[\mathbf{x}_{t}, \mathbf{h}_{t-1},\mathbf{c}_{t-1}]+\mathbf{b}_{o}\right)\\
    \tilde{\mathbf{c}}_{t} &=\tanh \left(\mathbf{W}_{c}[ \mathbf{x}_{t},\mathbf{h}_{t-1}]+\mathbf{b}_{c}\right) \\
    \mathbf{c}_{t} &=\mathbf{f}_{t} * \mathbf{c}_{t-1}+\mathbf{i}_{t} * \tilde{\mathbf{c}}_{t} \\ 
    \mathbf{h}_{t} &=\mathbf{o}_{t} * \tanh(\mathbf{c}_{t})
\end{align*}
Where $\mathbf{W}_{i}$, $\mathbf{W}_{f}$, $\mathbf{W}_{o}$, and $\mathbf{W}_{c}$ represent the weight matrix for the different gates, $*$ is the element-wise multiplication and $\sigma$ is the element-wise sigmoid function. We used the Bidirectional LSTM (Bi-LSTM) which considers the dependency of future and previous steps by concatenating the hidden states of the stacked forward and backward LSTMs respectively [17]. The overall hidden state $\mathbf{H}_{t}$ for the Bi-LSTM is given as $\overrightarrow{\mathbf{h}_{t}} \oplus \overleftarrow{\mathbf{h}_{t}}$.
\begin{figure}[!h]
  \centering
   \includegraphics[width=1\linewidth]{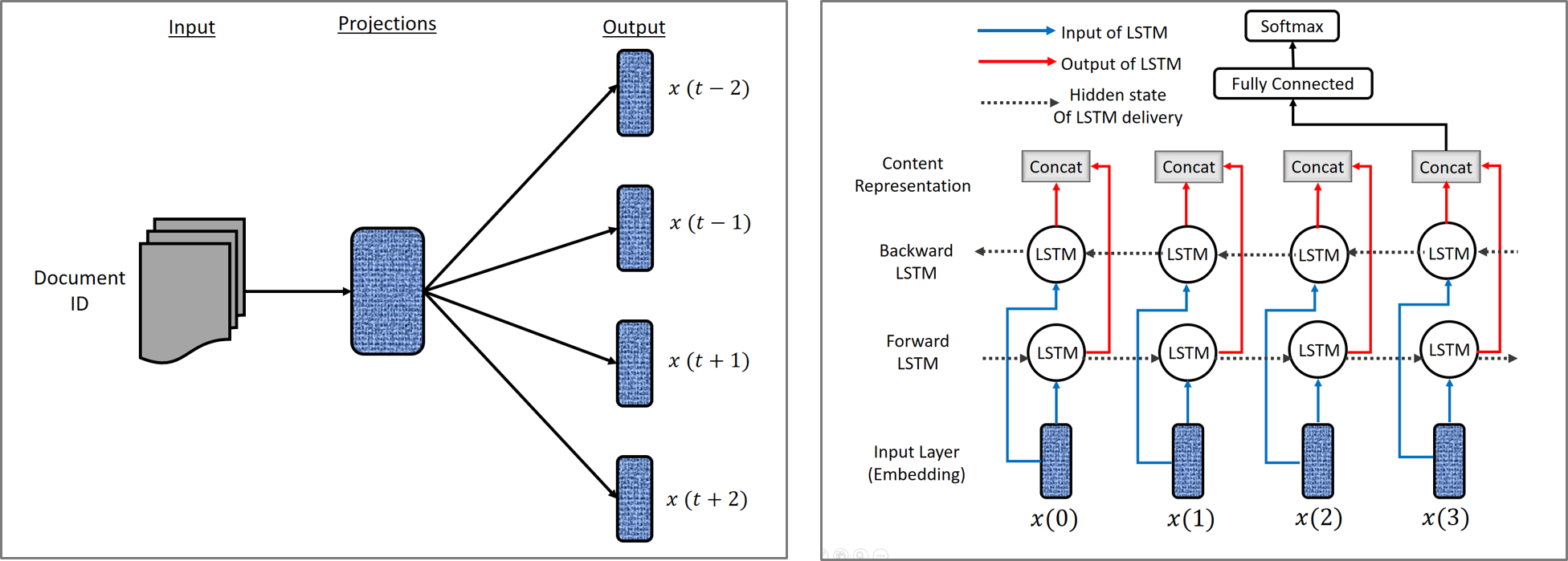}
  \caption*{Left to Right: A schematic of the PV-DBoW embedding and the Bi-LSTM architecture used in this project. }
\end{figure}
\newpage
\section*{Appendix B: Comparison between the different models}

% If you use beamer only pass "xcolor=table" option, i.e. \documentclass[xcolor=table]{beamer}
\begin{table}[!h]
\centering
\begin{tabular}{c|c|c|c}
\hline
\textbf{}                  & \textbf{Precision}                    & \textbf{Recall}                       & \textbf{F1-Score}                     \\ \hline
\textbf{BiLSTM + Doc2Vec}  & {\color[HTML]{9A0000} \textbf{0.732}} & {\color[HTML]{9A0000} \textbf{0.720}} & {\color[HTML]{9A0000} \textbf{0.718}} \\ \hline
\textbf{BiLSTM + Word2Vec} & 0.718                                 & 0.714                                 & 0.714                                 \\ \hline
\textbf{CNN + Doc2Vec}     & 0.711                                 & 0.699                                 & 0.698                                 \\ \hline
\textbf{MLP + Doc2Vec}     & 0.710                                 & 0.701                                 & 0.700                                 \\ \hline
\textbf{MLP + Word2Vec}    & 0.715                                 & 0.701                                 & 0.700                                 \\ \hline
\textbf{SVM + TFIDF}       & 0.722                                 & 0.710                                 & 0.709                                 \\ \hline
\end{tabular}
\caption*{Comparison of BiLSTM + Doc2Vec with other baseline methods}
\end{table}
\begin{figure}[!h]
  \centering
   \includegraphics[width=1\linewidth]{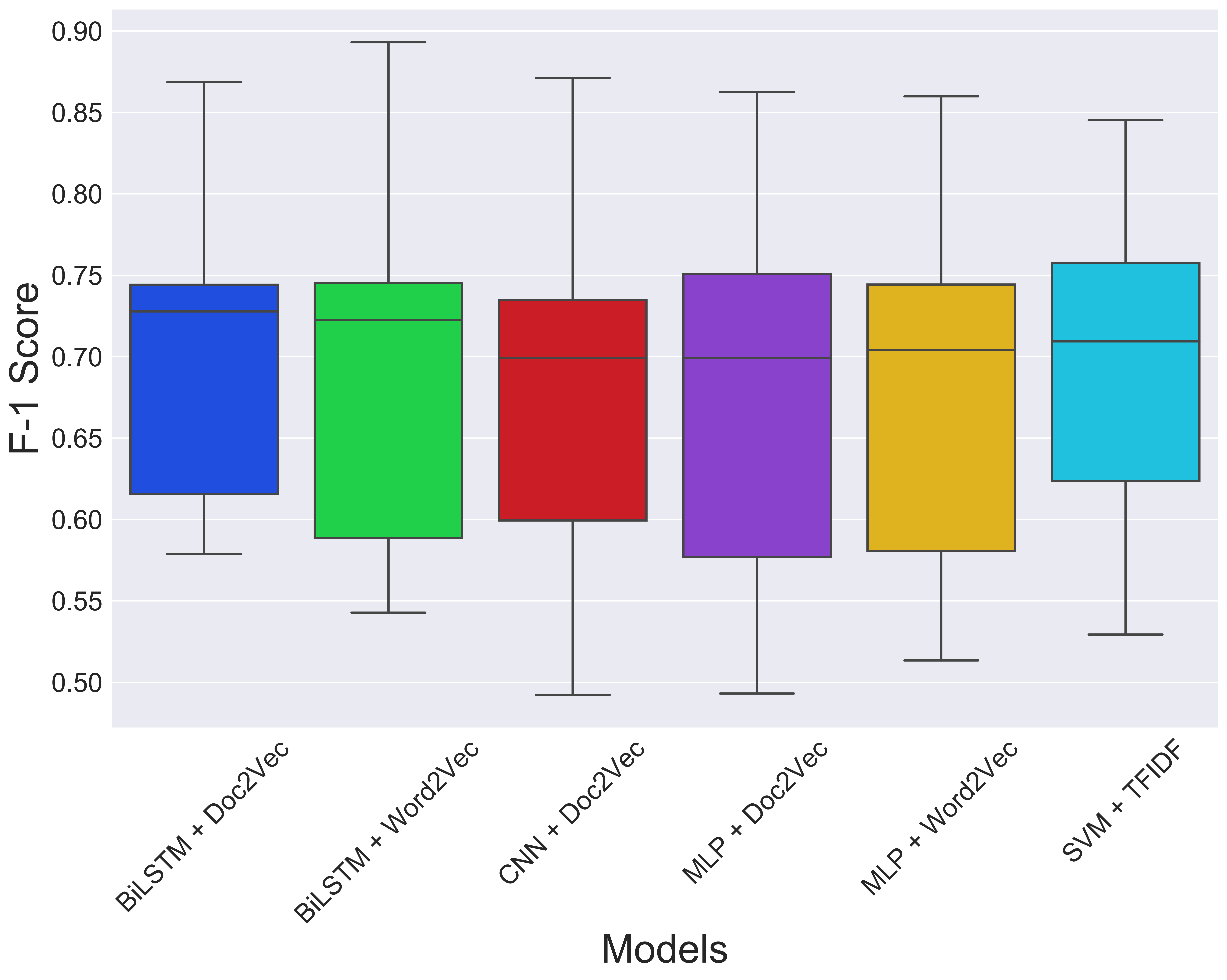}
  \caption*{Box Plot showing comparison with other models for the different classes. }
\end{figure}
\end{document}